\documentclass[journal]{IEEEtran}

\ifCLASSINFOpdf
\else
\fi

\usepackage[colorlinks,linkcolor=blue]{hyperref}
\usepackage{graphicx}
\usepackage[numbers]{natbib}

\begin{document}

\title{Weak Novel Categories without Tears: A Survey on Weak-Shot Learning}

\author{Li Niu
\thanks{Li Niu is with MOE Key Lab of Artificial Intelligence, Department of Computer Science and Engineering Shanghai Jiao Tong University, Shanghai, China (email: ustcnewly@sjtu.edu.cn).}
}

\maketitle

\begin{abstract}

Deep learning is a data-hungry approach, which requires massive training data. However, it is time-consuming and labor-intensive to collect abundant fully-annotated training data for all categories. Assuming the existence of base categories with adequate fully-annotated training samples, different paradigms requiring fewer training samples or weaker annotations for novel categories have attracted growing research interest. Among them, zero-shot (\emph{resp.}, few-shot) learning explores using zero (\emph{resp.}, a few) training samples for novel categories, which lowers the quantity requirement for novel categories. Instead, weak-shot learning lowers the quality requirement for novel categories. Specifically, sufficient training samples are collected for novel categories but they only have weak annotations. In different tasks, weak annotations are presented in different forms (\emph{e.g.}, noisy labels for image classification, image labels for object detection, bounding boxes for segmentation), similar to the definitions in weakly supervised learning. Therefore, weak-shot learning can also be treated as weakly supervised learning with auxiliary fully supervised categories. In this paper, we discuss the existing weak-shot learning methodologies in different tasks and summarize the codes at \href{https://github.com/bcmi/Awesome-Weak-Shot-Learning}{https://github.com/bcmi/Awesome-Weak-Shot-Learning}.
\end{abstract}

\IEEEpeerreviewmaketitle

\section{Introduction}\label{sec:intro}

Deep learning has achieved remarkable success in a variety of computer vision tasks including image classification \cite{krizhevsky2012imagenet,he2016deep}, object detection \cite{girshick2015fast,ren2015faster}, \emph{etc}. However, training deep learning models relies on massive fully-annotated training data from different categories, which are usually very expensive to acquire. There are a huge number of categories including fine-grained categories and new categories are also continuously emerging \cite{Meets,chen2020weak}. Thus, it is almost impossible to collect abundant fully-annotated training data for all categories. In practice, we generally have a set of base categories with adequate fully-annotated training samples. The problem is how to handle novel categories without adequate fully-annotated training samples, in which the novel categories have no overlap with base categories. 

Previous learning paradigms zero-shot learning~\cite{ZSL} and few-shot learning \cite{FSL} attempt to lower the quantity requirement for novel categories, that is, leveraging zero (\emph{resp.}, a few) training samples for novel categories. To bridge the gap between base categories and novel categories, zero-shot learning requires category-level semantic representation (\emph{e.g.}, word vector \cite{wordvector1} or human annotated attributes \cite{ZSL}) for all categories, while few-shot learning requires a few clean examples (\emph{e.g.}, $5$, $10$) for novel categories. Despite the great success of zero-shot learning and few-shot learning, they have the following drawbacks: 1) Annotating attributes or a few clean samples may require expert knowledge, which is not always available; 2) Word vector is free, but sometimes ambiguous (\emph{e.g.}, one word has multiple meanings) or unreasonable (\emph{e.g.}, average the word vectors of multiple words in one category name). Besides, word vector is much weaker than human annotated attributes~\cite{akata2015evaluation}.

\begin{figure}[t]
\begin{center}
\includegraphics[width=.95\linewidth]{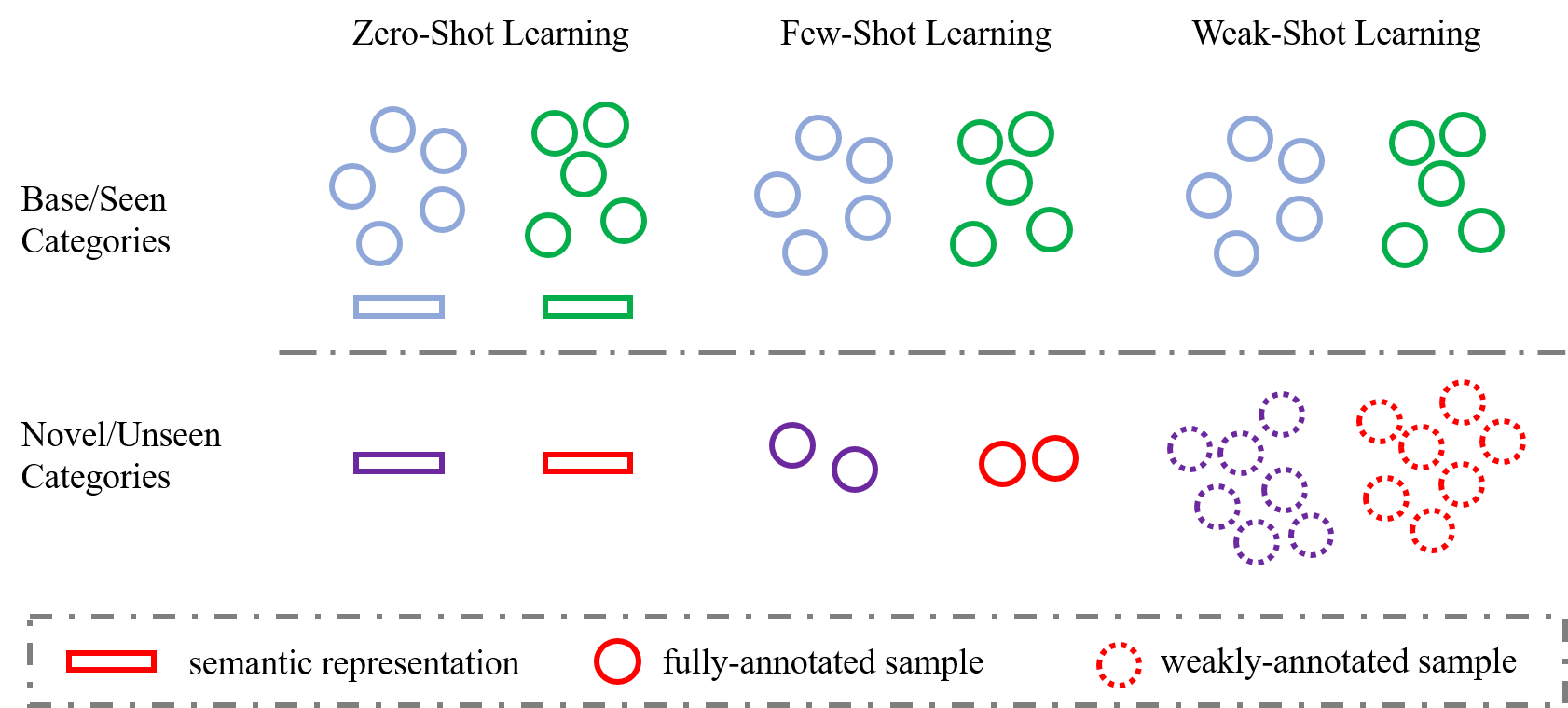}
\end{center}
\caption{ Comparison among zero-shot learning, few-shot learning, and weak-shot learning. Different colors indicate different categories.}
\label{fig:overview}
\end{figure}

Instead of lowering the quantity requirement for novel categories, we have an alternative choice, \emph{i.e.}, lowering the quality requirement for novel categories, resulting in another learning paradigm weak-shot learning \cite{chen2020weak}. In weak-shot learning, similar to zero/few-shot learning, all categories are split into base categories and novel categories without overlap. Base categories have abundant fully-annotated training samples while novel categories have abundant weakly-annotated training samples. The comparison among zero-shot learning, few-shot learning, and weak-shot learning is shown in Figure~\ref{fig:overview}. The key difference between weak-shot learning and zero/few-shot learning lies in weak or few training samples for novel categories.
The definitions of full annotation and weak annotation depend on different tasks, similar to weakly supervised learning. To name a few, for image classification task, full annotation indicates clean image label and weak annotation indicates noisy image label. For object detection task, full annotation indicates bounding box and weak annotation could be image label. For instance segmentation task, full annotation indicates instance mask and weak annotation could be image label or bounding box. For other tasks, full annotation and weak annotation could also be similarly defined as in weakly supervised learning paradigm. Therefore, weak-shot learning can also be treated as weakly supervised learning with auxiliary fully supervised categories. 

The key problem in weak-shot learning is transferring knowledge from base categories to novel categories to bridge the gap between weak annotation and full annotation on novel categories. Existing approaches can be roughly categorized into three groups: 1) transfer the category-invariant target; 2) transfer the mapping from weak annotation to full annotation; 3) decompose the entire task into a weakly-supervised subtask and a fully-supervised subtask. With knowledge transfer from base categories to novel categories, the performance on the novel categories could be significantly improved, compared with merely using weakly annotated training samples from novel categories. In the testing stage, the test samples could be only from novel categories, which is dubbed as weak-shot learning, or from both base categories and novel categories, which is dubbed as generalized weak-shot learning \cite{chen2020weak}. For ease of description, we refer to the training samples from base (\emph{resp.}, novel) categories as base (\emph{resp.}, novel) training samples. 
In the remainder of this paper, we will briefly review several related topics in Section~\ref{sec:related}. Then, we will describe the existing methodologies and applications of weak-shot learning in Section~\ref{sec:method} and Section~\ref{sec:task} respectively. Finally, we will conclude the whole paper in Section~\ref{sec:conclusion}.

\section{Related Works}\label{sec:related}
In this section, we briefly review several related topics: zero-shot learning, few-shot learning, weakly-supervised learning, and semi-supervised learning.

\subsection{Zero-Shot Learning}
Zero-shot learning employs category-level semantic representation (\emph{e.g.}, word vector or annotated attributes) to bridge the gap between seen (\emph{resp.}, base) categories and unseen (\emph{resp.}, novel) categories. A large part of works \cite{ZSL1,ZSL2,ZSL3,ZSL4} learn a mapping between visual features and category-level semantic representations. Recently, zero-shot object detection~\cite{zhu2019zero,demirel2018zero}, zero-shot semantic segmentation~\cite{bucher2019zero, gu2020context, xian2019semantic}, and zero-shot instance segmentation \cite{zheng2021zero} have also been explored. Zero-shot learning relies on category-level semantic representations for both base categories and novel categories, which are not required by weak-shot learning.

\subsection{Few-Shot Learning}
Few-shot learning depends on a few clean images (\emph{e.g.}, $5$-shot or $10$-shot) to learn each novel category, which could be roughly categorized as the following three types.
Optimization-based methods \cite{FSL_optim_1,FSL_optim_2} optimize the classifier on a variety of learning tasks (\emph{e.g.}, learn each category by a few images), such that it can solve new learning tasks using only a small number of images (\emph{e.g.}, learn novel categories with a few images).
Memory-based methods \cite{FSL_memory_1,FSL_memory_2,SNAIL} employ memory architectures to store key training images or directly encode fast adaptation algorithms.
Metric-based methods \cite{FSL_metric_1,FSL_metric_2,FSL_metric_3,MetaBaseline} learn a deep representation with a similarity metric in feature space and classify test images in a nearest neighbors manner. Recently, few-shot object detection~\cite{kang2019few,fan2020few}, few-shot semantic segmentation~\cite{fan2020fgn,zhang2020sg,dong2018few,hu2019attention,wang2019panet}, and few-shot instance segmentation~\cite{fan2020fgn} have also been explored. Few-shot learning relies on a few fully-annotated samples from novel categories, which are not required by weak-shot learning.

\subsection{Weakly-Supervised Learning}
Due to the data-hungry property of deep learning, learning from weakly annotated data has attracted increasing attention. Many weakly supervised classification methods have been proposed to deal with noisy images by outlier removal \cite{OutlierRemovel1,OutlierRemovel2}, robust loss function \cite{RobustLoss1,RobustLoss2,RobustLoss3}, label correction \cite{LabelCorrection1,LabelCorrection2,LabelCorrection3}, multiple instance learning \cite{MIL2,MIL3}, and so on \cite{zhang2020web,AdLoss,yao2019safeguarded,Unreasonable,niu2018learning}. Besides, weakly-supervised object detection \cite{bilen2016weakly}, weakly-supervised semantic segmentation~\cite{papandreou2015weakly, pathak2015constrained, pinheiro2015image}, and weakly-supervised instance segmentation~\cite{zhou2018weakly} have also been studied to bridge the gap between weak annotation and full annotation in different tasks. Weak-shot learning can be treated as weakly supervised learning with the assistance of auxiliary fully supervised base categories.

\subsection{Semi-Supervised Learning}
Semi-supervised learning~\cite{berthelot2019mixmatch,xie2020unsupervised} jointly utilizes a set of fully-annotated training samples and another set of unannotated or weakly-annotated training samples. Semi-supervised learning approaches usually impose prior regularization on weakly-annotated data or transfer knowledge from fully-annotated data to weakly-annotated data. Despite the similarity in the sense of transferring knowledge from fully-annotated data to weakly-annotated data, weak-shot learning involves cross-category knowledge transfer, which is much more challenging than semi-supervised learning.

\section{Methodologies}\label{sec:method}
Existing weak-shot learning methods can be roughly divided into the following three groups. The first group of methods transfer the category-invariant target (\emph{e.g.}, similarity, objectness, boundary, saliency, shape) from base categories to novel categories. The second group of methods transfer the mapping from weak annotation to full annotation across categories. The third group of methods decompose the entire task into a weakly-supervised subtask and a fully-supervised subtask. 

\subsection{Transfer the Category-Invariant Target}
Previous works have investigated several types of category-invariant targets (\emph{e.g.}, similarity, objectness, boundary, saliency, shape) which could be transferred from base categories to novel categories. 

\subsubsection{Similarity}
Semantic similarity, indicating whether two instances (\emph{e.g.}, image, proposal, pixel) belong to the same category, can be transferred from based categories to novel categories. 
The similarity predictor learnt on base categories can be applied to novel categories to predict pairwise similarities. The obtained pairwise similarities can be used to denoise the training instances or regularize feature learning for novel categories, which has been explored in classification~\cite{chen2020weak}, object detection~\cite{liuyan2021weak}, semantic segmentation~\cite{siyuan2021weak}, and instance segmentation~\cite{fan2020commonality}. 

\subsubsection{Objectness}
In object detection task, objectness indicates whether a proposal tightly encloses an object. The objectness predictor learnt on base categories can be applied to novel categories to localize category-agnostic objects. The universal object detector paves the way for learning object detectors for novel categories \cite{singh2018r,li2018mixed,zhong2020boosting,liuyan2021weak}. 

\subsubsection{Boundary}
In segmentation task, semantic boundaries separate the objects from different categories. The boundary predictor learnt on base categories can be applied to novel categories to predict semantic boundaries. The obtained boundaries can help identify unconfident pixels or segment objects for novel categories \cite{siyuan2021weak,fan2020commonality}. 

\subsubsection{Saliency}
In segmentation task, instance saliency could be transferred from base categories to novel categories to help segment objects for novel categories \cite{zhou2020learning}.

\subsubsection{Shape}
In segmentation task, the object shapes of novel categories might be analogous to those of base categories. Therefore, the object shapes of base categories could serve as a dictionary of shape priors to help infer the instance masks of novel categories~\cite{kuo2019shapemask}.

\subsection{Transfer the Mapping from Weak Annotation to Full Annotation}
For the fully-annotated base training samples, we usually have access to their weak annotations. For example, given the training images with annotated bounding boxes or segmentation masks, we can obtain their image labels. Therefore, we can acquire both weak annotations and full annotations for the base training samples, through which the mapping from weak annotation to full annotation can be learnt. Nevertheless, for weak-shot classification, the base training images are not associated with noisy image labels. Hence, the idea of transferring the mapping from weak annotation to full annotation may not be applicable to weak-shot classification. 

\subsubsection{Annotation Mapping}
Ideally, with the learnt mapping from weak annotation to full annotation, we can acquire the full annotations of novel categories based on their weak annotations. However, this goal is usually hard to achieve in practice. Thus, some methods make some twists when striving to achieve this goal. 
For example, the object detection method \cite{liuyan2021weak} first derives coarse semantic masks from weak annotations (\emph{i.e.}, image labels), followed by learning the mapping from coarse semantic masks and feature maps to full annotations (\emph{i.e.}, bounding boxes). Similarly, instance segmentation method \cite{biertimpel2021prior} also leverages the coarse masks derived from bounding box labels to help segment objects. 

\subsubsection{Weight Mapping}
In lieu of directly learning the mapping from weak annotation to full annotation, some methods learn the mapping from the model weights based on weak annotation to the model weights based on full annotation. The object detection methods \cite{hoffman2014lsda,kuen2019scaling} obtain the classifier weights based on weak annotations (\emph{i.e.}, image labels) and the detector weights based on full annotations (\emph{i.e.}, bounding boxes). Then, they learn the mapping from classifier weights to detector weights. Similarly, the  instance segmentation method \cite{hu2018learning} learns the mapping from box weights to mask weights. 

\subsection{Task Decomposition} \label{sec:task_decomposition}
Another scheme is decomposing the entire task into a weakly-supervised subtask and a fully-supervised subtask. The weakly-supervised subtask is supervised by weak annotations of both base and novel catgories, while the fully-supervised subtask is supervised by full annotations of only base categories. Moreover, the knowledge learnt in the fully-supervised subtask can be transferred across categories. 

For example, a weak-shot semantic segmentation approach SimFormer \cite{simFormer} was developed based on MaskFormer \cite{cheng2021per}, which decomposes semantic segmentation task into proposal classification subtask and proposal segmentation subtask. Specifically, MaskFormer~\cite{cheng2021per} produces some proposal embeddings from shared query embeddings for the input image. Each proposal embedding is assigned to one category present in the image (empty assignment is allowed), which is referred to as proposal classification subtask. Then, the semantic similarities between each proposal embedding and all pixel-level embeddings are calculated to produce the segmentation mask corresponding to the category of this proposal embedding, which is referred to as proposal segmentation subtask. The proposal classification subtask is supervised by image-level labels of both base categories and novel categories. In contrast, the proposal segmentation subtask can only be supervised by pixel-level labels of base categories. However, proposal segmentation essentially learns the semantic similarities between proposal embedding and pixel-level embeddings, which is tranferrable from base categories to novel categories. By virtue of transferred semantic similarities, the proposal embeddings of novel categories are able to produce the segmentation masks for novel categories.  

\section{Applications}\label{sec:task}

In this section, we introduce weak-shot image classification, object detection, semantic segmentation, and instance segmentation. Interestingly, weak-shot object detection and instance segmentation have attracted relatively more research interest.

\subsection{Weak-Shot Image Classification}
In weak-shot image classification, the base training samples have clean image labels while the novel training samples have noisy image labels, in which the novel training samples are usually web images crawled from public websites (\emph{e.g.}, Flickr, Google) by using category names as queries \cite{Meets,chen2020weak}. 

The first work on weak-shot image classification is SimTrans \cite{chen2020weak}. In \cite{chen2020weak}, similarity network is trained on base categories and applied to novel categories. The obtained pairwise similarities on novel categories are used in two ways. On the one hand, the average similarity of each image is used to weigh its classification loss. 
Specifically, for the noisy training images within a novel category, the non-outliers (images with the correct labels) are usually dominant, while outliers (images with incorrect labels) are from non-dominant inaccurate categories.  When calculating the semantic similarities among training images within a novel category, outliers are more likely to be dissimilar to most other images.  Therefore, whether an image is an outlier can be determined according to its average similarity to other images. The classification losses of outliers are assigned with smaller weights, which contributes to learning a more robust classifier.
On the other hand, the pairwise similarities form the graph structure to facilitate feature learning. 
Specifically, when directly learning on novel training samples, the feature graph structure (the similarities among image features) is determined by noisy labels.
The classification loss implicitly pulls features close for images with the same labels. However, the feature graph structure may be misled by noisy labels, so transferred similarities can rectify the misled feature graph structure by enforcing the features of semantically similar images to be close. 

\subsection{Weak-Shot Object Detection}
In weak-shot object detection, the base training samples have bounding boxes while the novel training samples only have image labels. Weak-shot object detection is also called mixed-supervised or cross-supervised object detection in previous literature \cite{chen2020cross,li2018mixed}. The first weak-shot object detection work can date back to LSDA \cite{hoffman2014lsda}. 

Due to the lack of box-level annotations, pure weakly supervised object detection on novel categories heavily relies on the quality of candidate proposals, and the learned detector can not refine the proposals. Thanks to fully-annotated base categories, weak-shot object detection methods~\cite{li2018mixed,kuen2019scaling,
zhong2020boosting,liuyan2021weak,rochan2015weakly,tang2016Large}, mainly focus on mining various types of knowledge in the base categories and transferring them to novel categories. 
For example, previous methods \cite{hoffman2014lsda,kuen2019scaling} transfer the difference between classifier and detector from base category to novel category. In \cite{li2018mixed}, category-invariant objectness is learnt from base categories with an adversarial domain classifier. With the help of learned objectness knowledge, this method can distinguish objects and distractors, thus improving the ability to reject distractors in novel categories. 
In \cite{zhong2020boosting}, they employ one-class universal detector learned from base category to provide proposals for the Multi-Instance Learning (MIL) classifier, which are used to mine pseudo ground-truth for iterative refinement.
One recent work in \cite{chen2020cross} exploit the spatial correlation between high-confidence bounding boxes output by the MIL classifier.
More recently, inheriting the objectness transfer and iterative refinement strategy from \cite{zhong2020boosting}, TraMaS \cite{liuyan2021weak} additionally explores two transfer targets: mask prior and semantic similarity. The transferred mask prior could help obtain better candidate bounding boxes, while the transferred semantic similarity could help discard the noisy pseudo bounding boxes.

\subsection{Weak-Shot Semantic Segmentation}
In weak-shot semantic segmentation, the base training samples have segmentation masks while the novel training samples only have image labels.

The first work on weak-shot semantic segmentation is RETAB \cite{siyuan2021weak}, which is designed under the typical weakly supervised segmentation framework and focuses on better expanding the initial response.  Two types of category-invariant information, \emph{i.e.}, semantic affinities and semantic boundaries, are transferred from base categories to novel categories. RETAB \cite{siyuan2021weak} contains an affinity learning step and an affinity-based propagation step.  In the affinity learning step, an affinity network  learns semantic affinities from ground-truth labels of base samples and CAMs~\cite{zhou2016learning} of novel samples simultaneously. In the affinity-based propagation step, a novel two-stage propagation strategy is proposed to propagate and revise CAMs by making full use of semantic affinities and semantic boundaries. More recently, SimFormer \cite{simFormer} accomplishes weak-shot semantic segmentation in a more compact and elegant framework. 
SimFormer \cite{simFormer} is built upon MaskFormer \cite{cheng2021per}, which decomposes semantic segmentation task into a proposal classification subtask and a proposal segmentation subtask. Besides the proposal-pixel similarity transfer in proposal segmentation subtask as mentioned in Section~\ref{sec:task_decomposition}, SimFormer \cite{simFormer} also transfers pixel-pixel similarity from base categories to novel categories for better performance.

\subsection{Weak-Shot Instance Segmentation}
In weak-shot instance segmentation, the base training samples have instance masks while the novel training samples only have bounding boxes. In contrast with with weak-shot object detection, weak-shot instance segmentation is consistently called partially-supervised instance segmentation in previous literature \cite{
hu2018learning,biertimpel2021prior,fan2020commonality}.

The first partially-supervised instance segmentation method is Mask$^{X}$ RCNN  \cite{hu2018learning}, which learns the mapping from box weights to mask weights. 
ShapeProp \cite{zhou2020learning} transfers instance saliency from base categories to novel categories. CPMask~\cite{fan2020commonality} transfers foreground boundaries as auxiliary information and transfers semantic affinity to regularize feature learning. OPMask~\cite{biertimpel2021prior} adds coarse mask prior derived from bounding box label to the mask head to predict more accurate instance mask.
ShapeMask \cite{kuo2019shapemask} transfers shape priors from base categories to novel categories. They first collect a set of shape priors from base categories. Given a detected box of novel category, the predicted object shape is calculated as the weighted sum of shape priors. 

\section{Conclusion}\label{sec:conclusion}

In this paper, we have introduced the formal definition of weak-shot learning. Zero/few-shot learning attempts to lower the quantity requirement for novel categories while weak-shot learning attempts to lower the quality requirement for novel categories. Similar to zero/few-shot learning, the key problem in weak-shot learning is also what and how to transfer from base categories to novel categories. We have also summarized the methodologies in existing weak-shot classification/detection/segmentation works, which will hopefully inspire the future works in this field.

\ifCLASSOPTIONcaptionsoff
  \newpage
\fi

\bibliographystyle{plainnat}
\bibliography{main.bbl}

\end{document}